\documentclass[amsa]{ipart}
\usepackage{amsfonts}
\usepackage{amsthm,amsmath}
\usepackage{paralist}
\usepackage{graphics} 
\usepackage{epsfig} 
\usepackage{graphicx} 
\usepackage{epstopdf}
\usepackage[colorlinks=true]{hyperref}
\usepackage{subfig}
\usepackage{array,multirow}
\usepackage{bm}
\usepackage{algorithm}

\allowdisplaybreaks
\newcommand\numberthis{\addtocounter{equation}{1}\tag{\theequation}}

\def\P{\mathbb{P}}
\def\N{\mathcal{N}}

\def\la{\left\langle}
\def\ra{\right\rangle}

\def\lb{\left(}
\def\rb{\right)}

\def\lcb{\left\{}
\def\rcb{\right\}}

\def\R{\mathbb{R}}

\DeclareMathOperator*{\divg}{div}
\DeclareMathOperator*{\diag}{diag}
\pubyear{2017}
\volume{0}
\issue{0}
\firstpage{1}
\lastpage{1}

\startlocaldefs
\endlocaldefs

\begin{document}

\begin{frontmatter}

\title{New region force for variational models in image segmentation and high dimensional data clustering }

\author{\fnms{Ke} \snm{Wei,}\thanksref{t1}\ead[label=e1]{kewei@math.ucdavis.edu}}
\thankstext{t1}{The work of this author was supported by the National Science Foundation under grant number DTRA-DMS 1322393.}
\address{Department of Mathematics\\ University of California at Davis\\California, USA\\\printead{e1}}
 \author{\fnms{Ke} \snm{Yin,}\thanksref{t2}\corref{}\ead[label=e2]{kyin@hust.edu.cn}}
\thankstext{t2}{Corresponding author.}
\address{Center for Mathematical Sciences\\Huazhong University of Science and Technology\\Wuhan, China\\\printead{e2}}
\author{\fnms{Xue-Cheng} \snm{Tai}\thanksref{t3}\ead[label=e3]{tai@math.uib.no}}
\thankstext{t3}{This author acknowledges the support from Norwegian Research Council through ISP-Matematikk (Project no. 239033/F20).}
\address{
Department of Mathematics, \\ University of Bergen, Postboks 7800, 5020, Norway. 
\\\printead{e3}}
\and
\author{\fnms{Tony F.} \snm{Chan}\thanksref{}\ead[label=e4]{tonyfchan@ust.hk}}
\address{Office of President\\ Hong Kong University of Science and Technology\\Hong Kong, China\\\printead{e4}}

\begin{abstract}
We propose an effective framework for multi-phase image segmentation and semi-supervised data clustering by introducing a novel region force term into the Potts model.  
Assume the probability that a pixel or a data point belongs to each class is known a priori. We show that the corresponding indicator function obeys the Bernoulli distribution and the new region force function can be computed as the negative log-likelihood function under the Bernoulli distribution. We solve the Potts model by the primal-dual hybrid gradient method and the augmented Lagrangian method, which are based on two different dual problems of the same primal problem. Empirical evaluations of the Potts model with the new region force function on benchmark problems show that it is competitive with  existing variational methods in both image segmentation and semi-supervised data clustering. 
\end{abstract}



\end{frontmatter}
 \section{Introduction}
Image segmentation plays an important role in image processing and appears in  a wide range of applications, including computer vision \cite{PaChFa050a}, stereo \cite{KoZa020a,KoZa040a} and 3D reconstruction \cite{VETC070a}. Given an image $I(x)$ defined over a domain $\Omega\in\R^2$, the task is to partition $\Omega$ into different subdomains so that $I(x)$ has different properties over each subdomain.  
According to different criteria, image segmentation can be divided into  two-phase segmentation vs. multi-phase segmentation, automatic segmentation vs. user-assisted segmentation, and discrete approach based segmentation vs. continuous approach based segmentation, just to name a few.
In this paper we study the multi-phase image segmentation problem where the number of partitions, denoted by $K$, is known a priori. 

In the spatially discrete setting, a digital image is usually modeled 
as a graph, and the solution to the multi-phase image segmentation problem can be computed from the min-cut or max-flow solutions of 
the graph, see \cite{BoKo05a,KoZa04a,BoVeZa03a,BoKo03a} and references therein. In contrast,  variational approaches
have been widely studied in the spatially continuous setting, where image segmentation is typically formulated as   a continuous energy-functional minimization problem over the image domain.  
We consider the Potts model for multi-phase image segmentation.  In the simplest form, the Potts model attempts to partition an image by minimizing an energy-functional which combines a region force term and an edge force term.
It has several advantages compared with the graph-based approaches: (i) it can avoid the metrication errors  owing to the crucial rotation invariance property; (ii) a wide range of reliable numerical algorithms are available, and those algorithms
can be easily implemented and accelerated; (iii) it requires less memory in computation; (iv) it is easy to use GPU and  other parallel computing systems. The active contours  which first appeared as the snake model in \cite{snakes1988} is another well known approach in variational models. In the active contours approach, the boundaries of each subdomain are modeled as curves, which can be evolved by minimizing an energy-functional.

Data clustering (or classification) is a fundamental task in machine learning which is about partitioning a large data set into a number of clusters that can be well interpreted from a practical perspective.  In general, 
data clustering can be roughly divided into three groups:  unsupervised clustering, supervised clustering and semi-supervised clustering. 
In this paper we consider the multi-class semi-supervised data clustering problem in which the number of clusters is given and there are a few  labelled data points  in each cluster.  The goal is to infer labels for the rest of data points from the already labelled ones. In practice, data points are typically modeled as vertices of a weighted graph where the weights on the edges describe the affinity between each pair of data points. Many algorithms have been developed under the graphical model. For example, the idea of geometric diffusion was developed for semi-supervised  clustering in a seminal paper by Coifman et al. \cite{coifman_geometric_2005} . The propagation of labels in geometric diffusion is driven by a diffusion kernel on the weighted  graph of the data points. 
Moreover, the diffusion map based on the eigenvectors of the graph Laplacian embeds the data points into a feature space with the diffusion distance as a new metric. Variational approaches have also been extended from image segmentation in spatially continuous domain to data clustering on weighted graphs, which will be the focus of this paper. In \cite{bresson_multi-class_2013}, the Mumford-Shah-Potts model \cite{MumfordShah1989, Potts1952} was demonstrated to be effective for data clustering, where the Cheeger cut, formulated as the sum of a modified total variation of the cluster indicator functions, is used.
The Cheeger cut  can be  interpreted as the perimeter of a cluster normalized by the imbalance of the cluster sizes and hence acts as the edge force in the model. In \cite{hu_multi-class_2015,lezoray_nonlocal_2012,garcia-cardona_multiclass_2014, merkurjev2013mbo}, the authors attempt to extend the Chan-Vese model \cite{chve01a} to data clustering, where the edge force is the total variation of the indicators functions. Furthermore, after approximating the total variation via the phase
field representations, a  graph-based Merriman-Bence-Osher (MBO \cite{merriman1992diffusion}) scheme  is developed to solve the diffusion equations with double-well potential. In all the aforementioned variational approaches, the region force term is  either based on the distance between the data points and the cluster centroids or based on the  mismatch of the labels  over the already labelled data.

The main contributions of this work are summarized in the following.

    (i) We  
    introduce a new region force term into the Potts model for both image segmentation and high dimensional data clustering. Compared with the snake model, a region force term was introduced for image segmentation in the Chan-Vese model \cite{chan_active_2001}. Some earlier works have tried to introduce region force into variational models for data clustering, see for example \cite{hu_multi-class_2015,lezoray_nonlocal_2012,Bae}. The region force introduced in \cite{yin2016newregion} overcomes some of the difficulties and shows good numerical performance.  In the present work, we derive a new region force function and show its applications to multi-phase image segmentation and semi-supervised data clustering.  
    
    (ii) The variational model we use in this paper is the Potts model. Using graph total variation, one can easily extend this model to high dimensional data clustering \cite{bresson2013multiclass}\cite{merkurjev2015global}. Following \cite[p.116]{bae2011global}, \cite[p.386]{yuan2010continuous} and \cite{yuan2014spatially,wei2015primal}, the Potts model has two different dual formulations. Related to these two dual formulations, we present two numerical algorithms. One is for the first dual formulation using a primal-dual algorithm, while the other one is for the second dual formulation using an augmented Lagrangian algorithm. 
    
    (iii) 
    Numerical experiments show the good performce of the new region force function and demonstrate the effectiveness of the numerical algorithms. 
    The tests for image segmentation show that the new region force function is as effective as the widely used $L_2$ fidelity in the literature. However, the $L_2$ fidelity or the Euclidean distance is not applicable for semi-supervised data clustering problems when there exists complex geometry within the data, while our region force function still works very well. The numerical results for semi-supervised data clustering show that our approach can achieve higher classification accuracy than other existing variational methods. Meanwhile, it is much easier to implement the numerical algorithms for our approach.

The remainder of this paper is organized as follows. The Potts model and the corresponding primal-dual formulations for image segmentation and data clustering  are presented in Sections~\ref{subsec:image} and \ref{subsec:clustering}.  The new region force function is introdcued in Section~\ref{subsub:region}.  In Section~\ref{sec:algs}, we present the numerical algorithms for the Potts model, and Section~\ref{sec:numerics} contains the numerical simulations. 
We conclude this paper in Section~\ref{sec:con} with  some additional remarks about future directions.

\section{Variational models and primal-dual formulations}\label{sec:primaldual}

For image segmentation, the snake models only consider edge force \cite{snakes1988}. The Chan-Vese model \cite{chan_active_2001} introduced a region force into variational image segmentation.  There were efforts to extend these region force for data clustering, \cite{lezoray_nonlocal_2012,hu_multi-class_2015}. However, these extensions have problems for data with complex geometries. In \cite{yin2016newregion,Bae}, the authors successfully introduced a region force for data clustering. In the following, we shall continue in this direction and will introduce a new region force for both image segmentation and data clustering. Moreover, we will combine them with  efficient  algorithms based on two different primal-dual formulations of the primal problem. 

To make the connections between image segmentation and data clustering clear, we first present the model for traditional image segmentation. Since data clustering can be formulated as a graph partitioning problem,  
 essentially the same mathematical model can be established for it based on the graph total variation. Therefore, the new region force function equally applies for image segmentation and data clustering.

\subsection{Multi-phase image segmentation}\label{subsec:image}
Let us start with image segmenation. Given a gray scale image function $I: \Omega \mapsto R $, the two-phase Chan-Vese \cite{chan_active_2001} model is trying to solve the following minimization problem: 
\begin{align*}
 \min_{\phi, c_1, c_2 }
& \lambda_1 \int_{\Omega} |I(x) - c_1|^2 H(\phi)   \mathrm{d}x 
+ \lambda_2 \int_{\Omega} |I(x) - c_2|^2 (1-H(\phi))  \mathrm{d}x  \\ 
& \qquad +\mu \int_\Omega |\nabla H(\phi)| \mathrm{d}x,
\end{align*}
where  (i) $\phi$ 
is a level set function whose zero level curves set represents
the segmentation boundary, (ii)  $H(\cdot)$ is the Heaviside function, (iii) $c_1$  and $c_2$ are two real numbers, and (iv)
$\lambda_1$ and $\lambda_2$ and $\mu$ are positive numbers. In this work, we shall use a more general model that extends the above mode, i.e., the so-called Potts model. 
The Potts model  for multi-phase image segmentation tries to minimize the following { energy-functional}:
\begin{align*}
\min_{\lcb\Omega_k\rcb_{k=1}^K}\sum_{k=1}^K\int_{\Omega_k}f_k(x) \mathrm{d}x +  R\lb \{\Omega_k\}_{k=1}^K\rb,\numberthis\label{eq:potts_model}
\end{align*}
where $\lcb\Omega_k\rcb_{k=1}^K$ is a partition of $\Omega$ such that $\cup_{k=1}^K\Omega_k=\Omega$ and $\Omega_k\cap\Omega_{k'}=\emptyset~\mbox{ for }~k\neq k'$. The integrand $f_k(x)$ in \eqref{eq:potts_model} is usually referred to as the {\em region force function} or the {\em fidelity term}. 
In case that  $f_k(x) = |I(x)-c_k|^2$,  we recover the Chan-Vese model \cite{chan_active_2001}  
or the piecewise constant Mumford-Shah model \cite{MumfordShah1989}. 
The regularization term $R\lb \{\Omega_k\}_{k=1}^K\rb$ measures the geometry properties of the boundaries of $\lcb\Omega_k\rcb_{k=1}^K$. The regularizer  used in this paper is the sum of the weighted length of each boundary, i.e., 
\begin{equation}
    R\lb \{\Omega_k\}_{k=1}^K\rb=\sum_{k=1}^K|\partial\Omega_k|_\alpha =\sum_{k=1}^K \int_{\partial\Omega_k} \alpha(x) ds, \label{eq:edge_detector}
\end{equation}
where $\alpha(x)\ge 0 $ is normally  called an edge detector. A popular choice for the edge detector is $\alpha(x) = \frac{\beta}{1 + \gamma |\nabla I_{\sigma}|^2 }$  with $\gamma$ and $\beta$ being some properly chosen constants and $I_{\sigma}$ is a Gaussian smoothing of the image function $I(x)$. In case $\alpha(x)=1$, the regularizer is the sum of the length of each boundary. 

Let $\phi_k(x) (1\leq k\leq K)$ be an indicator function associated with the $k$-th sub-domain,
\begin{align*}
\phi_k(x) = \begin{cases}
1 & x\in\Omega_k\\
0 & x\not\in\Omega_k.
\end{cases}
\end{align*}
It is true that 
$\int_{\partial\Omega_k}\alpha(x)ds =\int_\Omega \alpha(x) |\nabla\phi_k(x)| \mathrm{d}x$, so we can rewrite  
\eqref{eq:potts_model}  with the regularizer $R\lb \{\Omega_k\}_{k=1}^K\rb=\sum_{k=1}^K|\partial\Omega_k|_\alpha$ as\footnote{Throughout this paper we omit the independent variable notation $x$ when there is no risk of confusion, and we use $|\cdot|$, $|\cdot|_1$, and $|\cdot|_\infty$ to denote the $l_2$-norm, $l_1$-norm and $l_\infty$-norm, respectively. }
\begin{align*}
\min_{\substack{\phi_k\in\lcb0,1\rcb\\ [\phi_k]\in S}}\sum_{k=1}^K\int_{\Omega} f_k\phi_k \mathrm{d}x +  \sum_{k=1}^K\int_\Omega \alpha(x) |\nabla\phi_k| \mathrm{d}x,\numberthis\label{eq:potts_model_TV}
\end{align*}
where 
$$
[\phi_k]=(\phi_{1},\cdots,\phi_{K}),
$$
and
$$
S = \lcb [\phi_k]:~\sum_{k=1}^K\phi_k=1,~0\leq\phi_k\leq 1\rcb.
$$

One can immediately see that \eqref{eq:potts_model_TV} is a non-convex optimization problem. Therefore there does not exist a tractable way to compute its global solution reliably.  In a seminal paper, Chan, Esedoglu and Nicolova \cite{ChEsNi060a} proposed to relax the binary value constraint on $\phi_k$ to $0\leq\phi_k\leq 1$. 
Based on this relaxation, \eqref{eq:potts_model_TV} can be  transformed into the following convex programming:
\begin{gather}
\min_{\substack{0\leq\phi_k\leq 1\\ [\phi_k]\in S}}\sum_{k=1}^K\int_{\Omega} f_k\phi_k \mathrm{d}x +  \sum_{k=1}^K\int_\Omega \alpha(x) |\nabla\phi_k| \mathrm{d}x.\label{eq:potts_model_conv}\tag{P}
\end{gather}
We will refer to \eqref{eq:potts_model_conv} as the {\em primal problem}. Despite its convexity, numerical algorithms for the primal problem usually suffer from slow convergence rate due to  the non-smoothness of the TV term. In this paper, we will present two numerical methods for \eqref{eq:potts_model_conv}: a primal-dual hybrid gradient descent method and an augmented Lagrangian method. In order to describe those two methods, we first give two dual formulations of \eqref{eq:potts_model_conv}, one of which leads to the continuous max-flow approach studied in \cite{YBCB140a}.

The first dual formulation of \eqref{eq:potts_model_conv} can be obtained using the following equality: 
\begin{align*}
\int_\Omega\alpha(x) |\nabla\phi_k| \mathrm{d}x = \max_{|q_k|\leq \alpha(x) }\int_\Omega\phi_k\divg q_k \mathrm{d}x,~k=1,\cdots,n,\numberthis\label{eq:tv_dual}
\end{align*}
and  the min-max theorem \cite[Chapter 6, Proposition 2.4]{EkTe990a}. That is,  
\begin{align}
&\min_{\substack{0\leq\phi_k\leq 1\\ [\phi_k]\in S}}\sum_{k=1}^K\int_{\Omega} f_k\phi_k \mathrm{d}x +  \sum_{k=1}^K\int_\Omega \alpha(x) |\nabla\phi_k| \mathrm{d}x\nonumber \\
=&\min_{\substack{0\leq\phi_k\leq 1\\ [\phi_k]\in S}}\max_{|q_k|\leq\alpha(x) }\sum_{k=1}^K\int_{\Omega} f_k\phi_k \mathrm{d}x + \sum_{k=1}^K\int_\Omega\phi_k\divg q_k \mathrm{d}x\nonumber\\
=&\max_{|q_k|\leq\alpha (x) }\min_{\substack{0\leq\phi_k\leq 1\\ [\phi_k]\in S}}\sum_{k=1}^K\int_{\Omega}\phi_k(f_k+\divg q_k) \mathrm{d}x\nonumber\\
=&\max_{|q_k|\leq\alpha (x) }\int_\Omega\min_{k=1,\cdots,K}(f_k+\divg q_k) \mathrm{d}x,\label{eq:dual1}\tag{D1}
\end{align}
The above dual formulation for \eqref{eq:potts_model_conv} was first observed in \cite[p.116]{bae2011global}, where a gradient decent method was developed to solve the smoothed dual problem \cite[p.120]{bae2011global}. In this work, we shall use the recent developed primal-dual algorithms related to the ones in  \cite{EsZhFc100a,ChPo110a,zhu_efficient_2008} to solve it, see Algorithm 1.

The second dual formulation of \eqref{eq:potts_model_conv} is given by 
\begin{align}
 \max_{\lambda}\int_\Omega \lambda  \mathrm{d}x\quad\mbox{subject to}\quad
 \begin{cases}
 h_k\leq f_k, |q_k|\leq\alpha (x) \\
 \divg q_k-\lambda +h_k=0.
 \end{cases}\label{eq:dual3}\tag{D2}
\end{align}
In fact, the above problem  is  a continuous max-flow problem with flow conservation in a system where the image region is copied $K$ times. The equality $\divg q_k - \lambda + h_k=0$ represents flow conservation in each of the copied regions. The vector function  $q_k$ is the flow inside each copy and the scalar function $h_k$ is the flow between the copies with upper flow constraint $h_k \le f_k$, see \cite[p.386]{yuan2010continuous}. A more comprehensive exploration about the connection between other continuous min-cut and max-flow problems were also discussed in \cite{wei2015primal}.   
In order to derive the above max-flow model, we begin with introducing $K$ auxiliary variables $h_k, ~k=1,\cdots,K$, 
\begin{align*}
&\min_{\substack{0\leq\phi_k\leq 1\\ [\phi_k]\in S}}\sum_{k=1}^K\int_{\Omega} f_k\phi_k \mathrm{d}x +  \sum_{k=1}^K\int_\Omega \alpha(x) |\nabla\phi_k| \mathrm{d}x \\
=&\min_{\substack{0\leq\phi_k\leq 1\\ [\phi_k]\in S}}\max_{h_k\leq f_k}\sum_{k=1}^K\int_{\Omega} h_k\phi_k \mathrm{d}x +  \sum_{k=1}^K\int_\Omega \alpha(x) |\nabla\phi_k| \mathrm{d}x\\
=&\min_{\substack{\phi_k\in\R\\ [\phi_k]\in S}}\max_{h_k\leq f_k}\sum_{k=1}^K\int_{\Omega} h_k\phi_k \mathrm{d}x +  \sum_{k=1}^K\int_\Omega \alpha(x) |\nabla\phi_k| \mathrm{d}x.\numberthis\label{eq:minmax3}
\end{align*}
To show the last line, suppose there exists a $\phi_k<0$. We can then take the corresponding $h_k$ to be negative infinity so that the inner maximum problem can be arbitrarily large. However, this case can be excluded since a minimization over $\phi_k$ is followed.

By introducing another  variable $\lambda $ and utilizing \eqref{eq:tv_dual}, one can further see that \eqref{eq:minmax3} is equivalent to
\begin{align*}
\min_{\substack{\phi_k\in\R}}\max_{\substack{\lambda \\h_k\leq f_k\\|q_k|\leq\alpha(x)}}\int_\Omega( 1-\sum_{k=1}^K\phi_k)\lambda  \mathrm{d}x+\sum_{k=1}^K\int_{\Omega} h_k\phi_k \mathrm{d}x +  \sum_{k=1}^K\int_\Omega\phi_k\divg q_k \mathrm{d}x.\numberthis\label{eq:minmax2}
\end{align*}
At last, the application of  the min-max theorem implies that the optimal value of the above min-max problem is equal to the optimal value of the following  max-min problem 
\begin{align*}
&\max_{\substack{\lambda \\h_k\leq f_k\\|q_k|\leq\alpha(x)}}\min_{\substack{\phi_k\in\R}}\int_\Omega( 1-\sum_{k=1}^K\phi_k)\lambda  \mathrm{d}x+\sum_{k=1}^K\int_{\Omega} h_k\phi_k \mathrm{d}x +  \sum_{k=1}^K\int_\Omega\phi_k\divg q_k \mathrm{d}x\\
=&\max_{\substack{\lambda \\h_k\leq f_k\\|q_k|\leq\alpha (x) }}\min_{\substack{\phi_k\in \R}}\int \lambda  \mathrm{d}x+\sum_{k=1}^K\int_{\Omega}\phi_k(\divg q_k-\lambda +h_k) \mathrm{d}x\\
=&\max_{\substack{\lambda \\h_k\leq f_k\\|q_k|\leq\alpha(x) }}\int_\Omega\lambda\mathrm{d}x\quad\mbox{subject to}\quad\divg q_k-\lambda +h_k=0,
\end{align*}
where the last line gives the dual problem in \eqref{eq:dual3} after rearrangement. 

The pair of min-max problems in \eqref{eq:potts_model_conv} and \eqref{eq:dual3} are continuous analogue of the min-cut and max-flow problems in graph theory. In the discrete case, it is well-known that the min-cut problem is equivalent to the max-flow problem. The above analysis suggests this is also true in the spatially continuous setting. The interested reader can find more details about the continuous max-flow approaches in \cite{YBCB140a,BLT13a,BT150a} and references therein.

\subsection{Semi-supervised clustering} \label{subsec:clustering}
\subsubsection{Discrete Potts model for data clustering}\label{subsubsec:graph_model}
Before describing the discrete Potts model for data clustering, we briefly review some concepts related to the graphic model. Our description follows that in \cite{gilboa2008nonlocal} (also adopted in \cite{garcia-cardona_multiclass_2014} and numerous other works).
 In the graphic model data feature vectors are represented by vertices of a weighted graph $G=(V,E,w)$, where $V$ represents the set of vertices, $E$ represents the set of edges connecting different vertices, and $w$ represents the set of weights on the edges. The graph $G$ is typically sparse in real applications. For instance, in image segmentation each pixel is only connected with its four nearest neighbor pixels or pixels in a local image patch. In data clustering problems, data points are often assumed to be uniformly distributed on a low dimensional manifold endowed with a  Riemannian metric $d(\cdot,\cdot)$. Each data point on the manifold is usually connected with $s$-nearest neighbors for a small $s$, and together they form a local patch of the manifold. Therefore the graph $G$ can be constructed by $s$-Nearest-Neighbor ($s$-NN). In practice, the number of neighbor points $s$ may be determined by the dimension or co-dimension of the underlying manifold.

There are several  interesting weight functions in the literautre, for example the radial basis function (RBF \cite{scholkopf2004kernel})
\begin{align}\label{eq:rbf_kernel}
w(x_i,x_j) = \exp(-d(x_i,x_j)^2/(2\epsilon)),
\end{align}
and the Zelnik-Manor and Perona function (ZMP \cite{zelnik2004self})  
\begin{align}
w(x_i,x_j) = \exp\left(-d(x_i,x_j)^2/(\sigma(x_i)\sigma(x_j))\right),\label{eq:zmp_kernel}
\end{align}
where $\epsilon$  in \eqref{eq:rbf_kernel}  is a tuning parameter and $\sigma(\cdot)$ in \eqref{eq:zmp_kernel} measures the local
variance within the data. Another popular weight function in natural language processing is the cosine similarity function \cite{singhal2001modern}
\begin{align}w(x_i,x_j) = \cos(x_i,x_j)=\frac{\langle x_i,x_j \rangle}{|x_i||x_j|}.
\end{align}
Let $W=(w_{ij})$ be a weight matrix constructed from the weight function and $D=(d_{ii})$ be a diagonal matrix with the $i$-th diagonal entry being equal to the $l_1$-norm of the $i$-th row of $W$. The normalized affinity matrix defined via $\widehat{W}=D^{-1/2}WD^{-1/2}$ will be used later in the computation of the new region force  function.

We will introduce more nations in order to describe the discrete Potts model for data clustering. Let $N=|V|$, the total number of vertices of the graph $G$. For any $u\in L^2(V)$, the gradient of $u$ at the vertex $x_i$, denoted by $\nabla u(x_i)$, is defined as
\begin{align*}
\nabla u(x_{i})=(\partial_{x_{1}}u(x_{i}),\ldots,\partial_{x_{N}}u(x_{i})),
\end{align*}
where 
\begin{align*}
\partial_{x_{j}}u(x_{i})=w_{ij}(u(x_{j})-u(x_{i})). 
\end{align*}
Here we assume $w_{ij}=0$ and consequently $\partial_{x_{j}}u(x_{i})=0$ if $x_ix_j\not\in E$. For  any $q=(q(x_i)(x_j))\in L^{2}(V,L^2(V))$, the divergence of $q(x_i)$, denoted by $\divg q(x_i)$, is defined as
\begin{align*}
\divg q(x_{i})=\sum_{j=1}^{N}w_{ij}(q(x_{j})(x_{i})-q(x_{i})(x_{j})).
\end{align*}The computation of the divergence of $q(\cdot)$ over all the vertices of $G$ can be proceeded in the following matrix form
\begin{align*}
\divg q = (W\circ(q^{T}-q))\bm{1},
\end{align*}
where $\circ$ denotes the Hadamard product.
Moreover, one can easily verify that the divergence operator is the adjoint of the gradient operator which satisfies 
\begin{align*}
\la\nabla u,q\ra=\la u,\divg q\ra.\numberthis\label{eq:grad_div}
\end{align*}

Now we are ready to describe the discrete Potts model. Suppose 
we want to partition the data points into $K$ clusters, denoted by $V_1,\cdots,V_K$. 
If  the corresponding membership function $\phi_k(x_i)$ for the $k$-th cluster is  defined as
\begin{align*}
\phi_k(x_i) = \begin{cases}
1& \mbox{if } x_i\in V_k\\
0 & \mbox{otherwise},
\end{cases}\numberthis\label{eq:label_func_discrete}
\end{align*}
then the discrete counterpart of the Potts model in \eqref{eq:potts_model_TV} can be written as
\begin{align*}
\min_{\substack{\phi_k\in\lcb0,1\rcb\\ [\phi_k]\in S}}\sum_{k=1}^{K}\sum_{x_i\in V}f_k(x_i)\phi_{k}(x_i)+\sum_{k=1}^{K}\sum_{x_i\in V}\alpha(x_i)|\nabla \phi_k(x_i)|_1,\numberthis\label{eq:potts_discrete}
\end{align*}
where $f_k(\cdot)$ is a region force  function and $|\nabla \phi_k(x_i)|_1$ is  the anisotropic version of the total variation, 
\begin{align*}
\sum_{x_i\in V}\alpha(x_i)|\nabla \phi_{k}(x_i)|_1&=\sum_{x_i\in V}\sum_{x_j\in V}\alpha(x_i)w_{ij}|\phi_{k}(x_{j})-\phi_{k}(x_{i})|\\
&=|\diag(\alpha)W\diag(\phi_{k})-\diag(\alpha)\diag(\phi_{k})W|_{1}.
\end{align*}
In addition, one also has
\begin{align*}
\sum_{x_i\in V}\alpha(x_i)|\nabla \phi_{k}(x_i)|_1 
&=\sum_{x_i\in V}\max_{|q_k(x_i)|_\infty\leq \alpha(x_i)}\la \nabla \phi_{k}(x_i), q_k(x_i)\ra\\
&=\max_{|q_k|_\infty\leq \alpha(x_i)}\la\nabla\phi_k,q_k\ra\\
&=\max_{|q_k|_\infty\leq \alpha(x_i)}\la\phi_k,\divg q_k\ra\\
&=\max_{|q_k|_\infty\leq \alpha(x_i)}\sum_{x_i\in V} \phi_k(x_i)\divg q_k(x_i),\numberthis\label{eq:tv_dual_discrete}
\end{align*}
where in the first line we use the fact that $\ell_\infty$-norm is the dual norm of $\ell_1$-norm, and in the fourth line we apply \eqref{eq:grad_div}. 

It is evident that \eqref{eq:potts_discrete} is a non-convex problem and the application of the same convex relaxation technique as in \eqref{eq:potts_model_conv} leads to the following primal problem of the Potts model for data clustering
\begin{align}
\min_{\substack{\phi_k\in[0,1]\\ [\phi_k]\in S}}\sum_{k=1}^{K}\sum_{x_i\in V}f_k(x_i)\phi_{k}(x_i)+\sum_{k=1}^{K}\sum_{x_i\in V}\alpha(x_i)|\nabla \phi_k(x_i)|_1.\label{eq:potts_discrete_conv}\tag{\={P}}
\end{align}
Using a variant of \eqref{eq:tv_dual_discrete} and the same min-max argument as in Section~\ref{subsec:image}, we are also able to obtain two different dual formulations for \eqref{eq:potts_discrete_conv}, which are listed below:
\begin{align}
&\max_{|q_k|_1\leq\alpha(x_i)}\sum_{x_i\in V}\min_k(f_k(x_i)+\divg q_k(x_i))\label{eq:db1}\tag{\={D}1},\\
&\max\sum_{x_i\in V}\lambda(x_i)\quad\mbox{subject to}\quad
 \begin{cases}
 h_k(x_i)\leq f_k(x_i), |q_k(x_i)|_\infty\leq\alpha(x_i)\\
 \divg q_k(x_i)-\lambda(x_i) +h_k(x_i)=0.
 \end{cases}\label{eq:db3}\tag{\={D}2}
\end{align} 

In this paper, we investigate the discrete Potts model for semi-supervised data clustering. Suppose there exists a small fraction $S_k\subset V_k$ in each cluster such that  the label of the data points in $S_k$ is given. The goal is to determine the labels for the rest of the data points from those labelled ones. In our approach, the labelled data points will be used to compute the probabilities in the new region force function presented in the next section.
\subsubsection{Effective region force under the Bernoulli model}\label{subsub:region}
In \cite{hu_multi-class_2015}, Hu, Sunu and Bertozzi extended the Chan-Vese model to data clustering by combing a region force function of the form $f_k(x_i)=|x_i-c_k|^2$  with a special type of edge force function. Here $c_k$ denotes the centroid of the $k$-th cluster which can be computed as the weighted average of the data points in each cluster. The quadratic region force function defined using the Euclidean distance between the data points and the cluster centroids penalizes the heterogeneity of the data points within each cluster. It is effective for the Gaussian mixture model, where the data points are homogeneous in a visually smooth region. However, for many data clustering problems, there exists complex geometry within the data points and typically the clusters cannot be distinguished by the centroid of each cluster, for example in the three-circles synthetic data set. In this section, we present a different region force function which can be obtained as the negative log-likelihood function   under the Bernoulli model.

Let $p_k(x_i)$ denote the probability of a given data point $x_i$ belonging to the $k$-th cluster $V_k$. If $p_k(x_i)$ is known a priori,  
then the binary value label function $\phi_k(x_i)$ defined in \eqref{eq:label_func_discrete}  is a  random variable which satisfies the Bernoulli distribution and 
\begin{align*}
\P\{\phi_k(x_i) \} &= \begin{cases}
p_k(x_i)  & \mbox{if } \phi_k(x_i) = 1\\
1- p_k(x_i) & \mbox{if } \phi_k(x_i) = 0
\end{cases}\\
&= (p_k(x_i))^{\phi_k(x_i)}(1-p_k(x_i))^{1-\phi_k(x_i)}.
\end{align*}
Therefore, the negative log-likelihood function over all the data points is given by 
\begin{align*}
-\sum_{x_i\in V}\log (\P\{\phi_k(x_i) \}) &= \sum_{x_i\in V}\{-\log(p_k(x_i))\phi_k(x_i)+\log(1-p_k(x_i))\phi_k(x_i)\} \\
&\quad+ Const.,
\end{align*}
where the last term is a constant as we assume $p_k(x_i)$ is given. Without any prior information imposed on $\phi_k(x_i)$, we are interested in a realization which minimizes the negative log-likelihood function under the constraint $\sum_{k=1}^K\phi_k(x_i)=1$ for all $x_i$.
This can be achieved by computing the solution of the following minimization problem: 
\begin{align*}
&\min_{\phi_k(x_i)}\sum_{k=1}^K\sum_{x_i\in V}\{-\log(p_k(x_i))\phi_k(x_i)+\log(1-p_k(x_i))\phi_k(x_i)\} \\
&\mbox{s. t. }\phi_k(x_i)\in\{0,1\}\mbox{ and }\sum_{k=1}^K\phi_k(x_i) = 1 \mbox{ for all }i=1,\cdots, n.
\end{align*}

The above minimization problem provides us a new region force function for the Potts model. That is, we can set  
\begin{align*}
f_k(x_i)=-\log(p_k(x_i))+\log(1-p_k(x_i))\numberthis\label{eq:region_force1}
\end{align*}
 so that $$f_k(x_i)\phi_k(x_i)=-\log(p_k(x_i))\phi_k(x_i)+\log(1-p_k(x_i))\phi_k(x_i)$$ in \eqref{eq:potts_discrete}. 
Since $\log(t)\leq t-1$ for all $t>0$, we have 
\begin{align*}
&-\log(p_k(x_i))+\log(1-p_k(x_i))\leq \frac{1-2p_k(x_i)}{p_k(x_i)}.\numberthis\label{eq:region_force1_bd}
\end{align*}
The numerator in \eqref{eq:region_force1_bd} gives the region force function proposed in \cite{yin2016newregion},
\begin{align*}
f_k(x_i) =1-2p_k(x_i) \numberthis\label{eq:region_force2}.
\end{align*} 
Numerical simulations in Section~\ref{sec:numerics} demonstrate that the  region force functions in \eqref{eq:region_force1} and \eqref{eq:region_force2} are equally effective when used in the  Potts model for  image segmentation and semi-supervised clustering.
\subsubsection{Compute the probability}
We now describe how to compute $p_k(x_i)$, the probability that $x_i$ belongs to $V_k$, in the region force function. The idea behind the computation is simple. If a data point is ``much closer'' to the labelled data points in a cluster $V_k$ (i.e., the data points in $S_k$), then with high probability this data point should belong to the $k$-th cluster. So  $p_k(x_i)$ should be proportional  to the ``closeness'' between $x_i$ and $S_k$. Similar ideas can be found in \cite{wu2014multi,wu2013markov}, where a novel  learning algorithm based on the random Markov chain model was proposed for semi-supervised clustering. 

Recall from Section \ref{subsubsec:graph_model} that $\widehat{W}$ is the  normalized affinity matrix for the data points. Let $\widehat{W}^m=(\widehat{w}^{(m)}_{ij})$ be the $m$-th power of $\widehat{W}$. In \cite{coifman_geometric_2005}, the $m$-th diffusion distance between two data points $x_i$ and $x_j$ is defined as 
\begin{align*}
d^{(m)}(x_i,x_j) = \widehat{w}^{(m)}_{ii}+\widehat{w}^{(m)}_{jj}-2\widehat{w}^{(m)}_{ij},
\end{align*}
where $\widehat{w}^{(m)}_{ii}$ (and $\widehat{w}^{(m)}_{jj}$) describes the probability that a random walk starting from $x_i$ (and $x_j$) returns back to  $x_i$ (and $x_j$) after $m$ steps, and $\widehat{w}^{(m)}_{ij}$ describes the probability that a random walk starting from $x_i$ arrives at $x_j$ after $m$ steps. Thus, $\widehat{w}^{(m)}_{ij}$ measures the closeness between two data points. Based on the diffusion distance, we compute $p_k(x_i)$ in semi-supervised clustering as follows:
\begin{equation}\label{eq:region_probability}
p_{k}(x_{i})=\frac{\frac{1}{|S_{k}|}\sum _{j\in S_{k}}r_{ij}}{\sum_{k'=1}^{K}\frac{1}{|S_{k'}|}\sum_{j\in S_{k'}}r_{ij}},
\end{equation}
where 
\begin{align*}
r_{ij}={(\widehat{w}_{ij}^{(m)})^{2}}/({\widehat{w}_{ii}^{(m)}\widehat{w}_{jj}^{(m)}}),
\end{align*}
and with a slight abuse of notation $|\cdot|$ denotes the cardinality of a finite set. 
In the numerical simulations, we take $m=1\mbox{ or }2$ and set $p_k(x_i)=1/K$ when the denominator in \eqref{eq:region_probability} is zero.

The  region force  functions presented in Section~\ref{subsub:region} are also applicable for the multi-phase image segmentation problem. That is, we can take $f_k(x)$ to be either 
$
-\log(p_k(x))+\log(1-p_k(x))
$
or
$
1-2p_k(x)
$
 in \eqref{eq:potts_model_TV} and \eqref{eq:potts_model_conv}. Assume the image density of each subdomain obeys the  Gaussian random model given by $I(x)\sim \N(c_k,\sigma^2)$. 
  For each pixel $x$ in the image domain, the probability of $x$ belonging to the  $k$-th subdomain, denoted by $p_k(x)$, should be proportional to $\exp(-|I(x)-c_k|/2\sigma^2)$. Therefore we can compute $p_k(x)$ as follows:
 \begin{align*}
 p_k(x) = \frac{\exp(-|I(x)-c_k|/2\sigma^2)}{\sum_{k'=1}^K\exp(-|I(x)-c_{k'}|/2\sigma^2)}.\numberthis\label{eq:image_prob}
 \end{align*}

\section{Algorithms}\label{sec:algs}
In this section, we present  two numerical algorithms for computing the solutions to the primal problems \eqref{eq:potts_model_conv} and \eqref{eq:potts_discrete_conv}. Since \eqref{eq:potts_discrete_conv} is just a discrete version of \eqref{eq:potts_model_conv}, we only present the algorithms for \eqref{eq:potts_model_conv} but note one can easily extend them for \eqref{eq:potts_discrete_conv}. As stated previously, computing the solution to  \eqref{eq:potts_model_conv} directly suffers from  the non-smoothness of the TV term. Alternatively, we solve \eqref{eq:potts_model_conv} by  the primal-dual hybrid gradient  method  and the alternating direction method of multipliers (ADMM) which are targeting the min-cut problem \eqref{eq:potts_model_conv} and the max-flow problem \eqref{eq:dual3}, respectively.

In Section~\ref{sec:primaldual}, two dual problems are presented for the primal problem \eqref{eq:potts_model_conv}. If $[\phi_k^*]$ is the optimal solution of the primal problem and $[q_k^*]$ is the optimal solution of the first dual problem, then $([\phi_k^*],[q_k^*])$ forms a saddle point of the min-max problem
\begin{align*}\min_{\substack{0\leq\phi_k\leq 1\\ [\phi_k]\in S}}\max_{|q_k|\leq\alpha(x)}\sum_{k=1}^K\int_{\Omega}f_k\phi_k \mathrm{d}x + \sum_{k=1}^K\int_\Omega\phi_k\divg q_k \mathrm{d}x.
\end{align*}
A primal-dual hybrid gradient  algorithm can be developed for the above min-max problem, see Algorithms~\ref{alg:pdhg}. 
In each iteration, the primal and dual variables are updated successively by a projected gradient descent step, followed by an acceleration using the Nesterov's memory technique.
Algorithm~\ref{alg:pdhg} is a special case of the general primal-dual algorithms that have been well studied  in the literature. The convergence analysis of the primal-dual algorithms can be found in \cite{EsZhFc100a,ChPo110a,zhu_efficient_2008,bonettini_convergence_2012}.

\begin{algorithm}[!ht]
{\normalsize\begin{enumerate}
\item Update the dual variables $[q_k]$ by
\[
q^{l+1}_{k}=\Pi_{|q_k|\leq\alpha(x)}(q_{k}^l-\beta_{k}\nabla\phi_{k}^l).
\]
\item Update the primal variables $[\phi_k]$ by
\[
[\phi_{k}^{l+1}]=\Pi_{S}[\phi_{k}^l-\gamma_{k}(\divg q_{k}^l+f_{k})].
\]
\item Combine two adjacent steps
\[
[\phi_{k}^{l+1}]=\theta[\phi_{k}^l]+(1-\theta)[\phi_{k}^{l+1}],
\]
where we choose $\theta=-0.5$.
\end{enumerate}}
\caption{\label{alg:pdhg}Primal-Dual Hybrid Gradient (PDHG)
}
\end{algorithm}

As noted below the second dual formation \eqref{eq:dual3}, the continuous Potts model   \eqref{eq:potts_model_conv} can be interpreted as a continuous min-cut problem, while the corresponding max-flow problem is given by its dual formulation in \eqref{eq:dual3}. Therefore we can instead solve the dual problem by the ADMM algorithm. First note that the augmented Lagrangian associated with \eqref{eq:dual3} is 
\begin{align*}
&L(\lambda,[h_k],[q_k],[\phi_k])
=
\int_\Omega \lambda  \mathrm{d}x 
\\ &
+ \sum_{k=1}^K\int_\Omega\phi_k(\divg q_k-\lambda +h_k) \mathrm{d}x-\frac{c}{2}\sum_{k=1}^K\int_\Omega (\divg q_k-\lambda +h_k)^2 \mathrm{d}x.\numberthis\label{eq:Lag}
\end{align*}
Here we use $[\phi_k]$ to denote the  Lagrangian multipliers because when we use ADMM to solve the dual problem based on the augmented Lagrangian, the Lagrangian multipliers converge to the primal optimal solution.
The ADMM algorithm for the  augmented Lagrangian functional \eqref{eq:Lag} is presented in Algorithm~\ref{alg:admm}. In the algorithm, each dual variable is updated by solving a minimization subproblem when the other variables are fixed, and  the Lagrangian multiplies are updated by a gradient descent step. While the closed-form solutions to the minimization problems with respect to $\lambda$ and $[h_k]$ can be computed easily, the minimization problem with respect to $[q_k]$ does not have an explicit solution. However, we can compute its solution approximately using one step projected gradient descent.

\begin{algorithm}[!ht]
{\normalsize\begin{enumerate}
\item 
Update the dual variables
\begin{enumerate}
\item Update $\lambda$ by
\begin{align*}\lambda^{l+1}&=\underset{\lambda}{\arg\max}\int_{\Omega}\lambda\mathrm{d}x-\frac{c}{2}\sum_{k=1}^{K}\int_\Omega(\divg q^l_{k}-\lambda+h^l_k-\frac{\phi^l_{k}}{c})^{2}\,\mathrm{d}x\\
&=\frac{1}{K}\sum_{k=1}^{K}(\divg q_{k}^l+h_k^l-\frac{\phi^l_k}{c})+\frac{1}{Kc}.
\end{align*}
 \item
Update $[h_k]$ by
 \begin{align*}h_{k}^{l+1}&=\underset{h_{k}\leq f_{k}}{\arg\max}-\int_{\Omega}(\mathrm{div}q^l_{k}-\lambda^{l+1}+h_k-\frac{\phi^l_{k}}{c})^{2}\,\mathrm{d}x\\
 &=\min\{\frac{\phi^l_{k}}{c}+\lambda^{l+1}-\mathrm{div}q^l_{k},f_{k}\}.
 \end{align*}
\item Update $[q_k]$ by
\begin{align*}q_{k}^{l+1}=\underset{|q_{k}|\leq\alpha(x)}{\arg\max}-\int_{\Omega}(\mathrm{div}q_{k}-\lambda^{l+1}+h_k^{l+1}-\frac{\phi_{k}^l}{c})^{2}\,\mathrm{d}x,
\end{align*} which can be approximately solved by one step of projected gradient descend,
\begin{align*}
q_k^{l+1} = \Pi_{|q_k|\leq\alpha(x)}(q_k^l+\beta_l\nabla(\divg q^l_{k}-\lambda^{l+1}+h_k-\frac{\phi^l_{k}}{c})).
\end{align*}
\end{enumerate}
\item Update the Lagrangian multipliers
\begin{align*}
\phi_{k}^{l+1}=\phi_{k}^l-c(h_{k}^{l+1}+\divg q_{k}^{l+1}-\lambda^{l+1}).
\end{align*}
\end{enumerate}}
\caption{\label{alg:admm}ADMM for Augmented Lagrangian (ADMM)}

\end{algorithm}
\section{Numerical Experiments}\label{sec:numerics}
In this section, we  explore the performance of the Potts model \eqref{eq:potts_model_conv} and \eqref{eq:potts_discrete_conv} with the new region force function for multi-phase image segmentation and semi-supervised data clustering, and test the efficiency of the numerical algorithms presented in Section~\ref{sec:algs}.  
Both  PDHG (Algorithm~\ref{alg:pdhg}) and ADMM (Algorithm~\ref{alg:admm}) are implemented in MATLAB\textregistered \, and executed on a laptop. 

\subsection{Multi-phase image segmentation}\label{sec:num_image}
\begin{figure*}[!ht]
\centering
\subfloat[]{\includegraphics[width=0.5\textwidth]{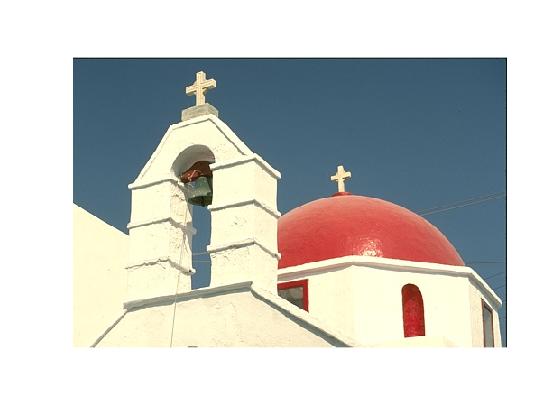}\label{chapel:orig}} \hfil
\subfloat[]{\includegraphics[width=0.5\textwidth]{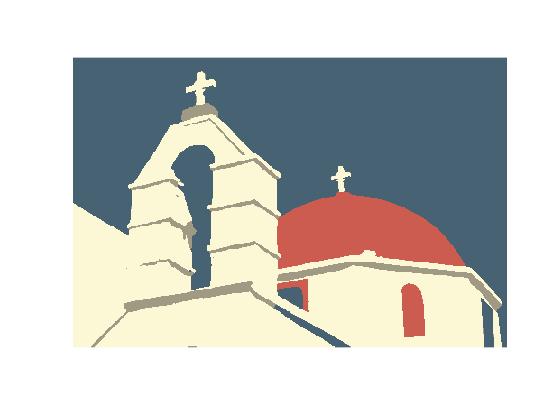}\label{chapel:logp}}\\
\subfloat[]{\includegraphics[width=0.5\textwidth]{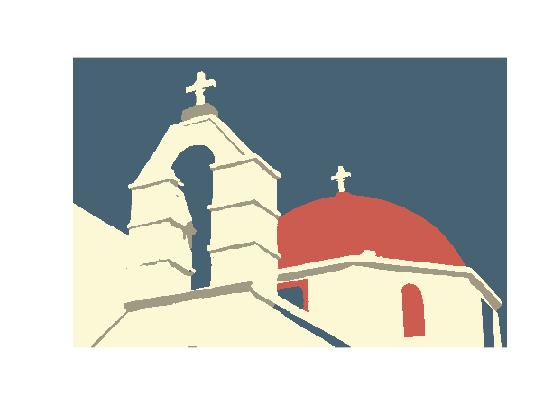}\label{chapel:p}}\hfil
\subfloat[]{\includegraphics[width=0.5\textwidth]{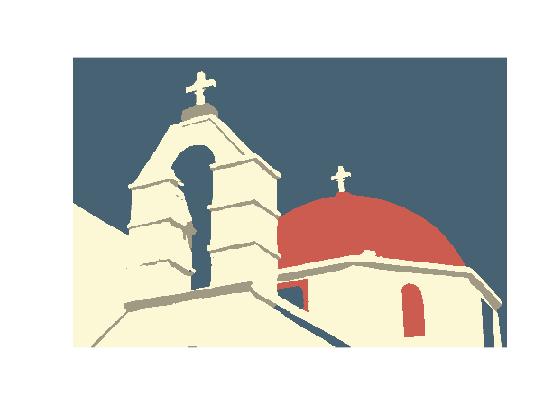}\label{chapel:l2}}
\caption{(a): Original image;  (b), (c), (d): Segmentation results obtained from the  Potts model with the region force functions~ \eqref{eq:region_force1},  \eqref{eq:region_force2}, and  \eqref{eq:region_L2}, respectively.}
\label{fig:chapel}
\end{figure*}

\begin{figure*}[!ht]
\centering
\subfloat[]{\includegraphics[width=0.4\textwidth,height=0.5\textwidth]{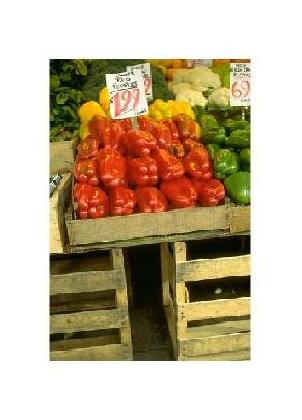}\label{veg:orig}} 
\subfloat[]{\includegraphics[width=0.4\textwidth,height=0.5\textwidth]{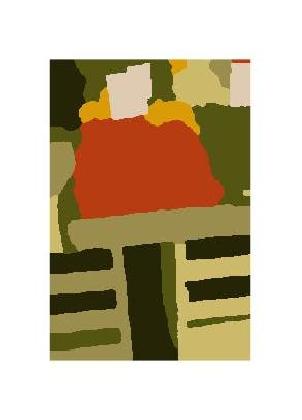}\label{veg:logp}}\\
\subfloat[]{\includegraphics[width=0.4\textwidth,height=0.5\textwidth]{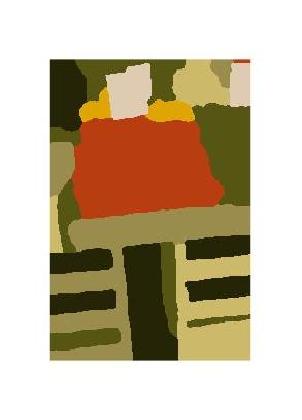}\label{veg:p}}
\subfloat[]{\includegraphics[width=0.4\textwidth,height=0.5\textwidth]{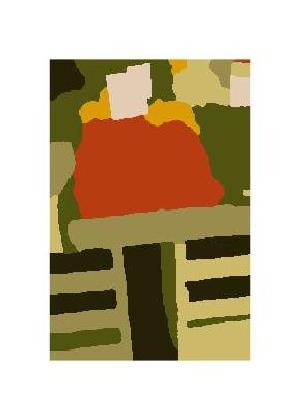}\label{veg:l2}}
\caption{(a): Original image;  (b), (c), (d): Segmentation results obtained from the  Potts model with the region force functions~ \eqref{eq:region_force1},  \eqref{eq:region_force2}, and  \eqref{eq:region_L2}, respectively.}
\label{fig:veg}
\end{figure*}

\begin{figure*}[!ht]
\centering
\subfloat[]{\includegraphics[width=0.5\textwidth]{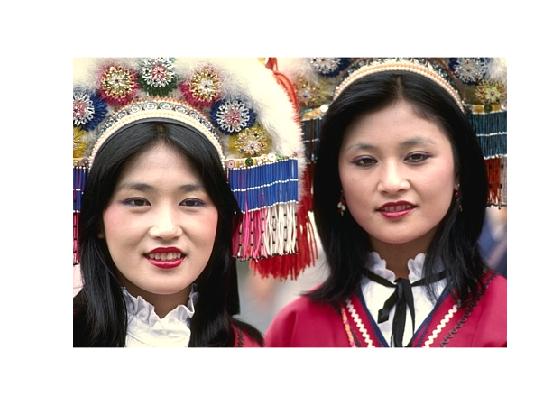}\label{lady:orig}} \hfil
\subfloat[]{\includegraphics[width=0.5\textwidth]{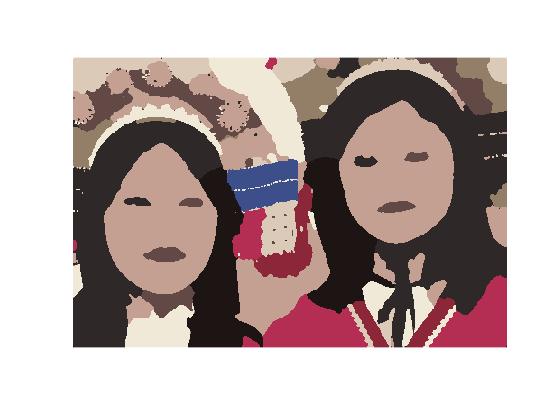}\label{lady:logp}}\\
\subfloat[]{\includegraphics[width=0.5\textwidth]{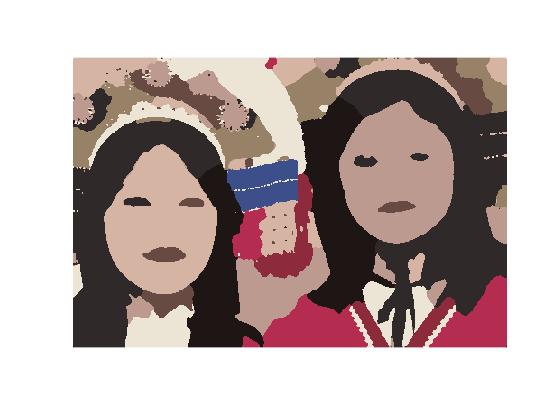}\label{lady:p}}\hfil
\subfloat[]{\includegraphics[width=0.5\textwidth]{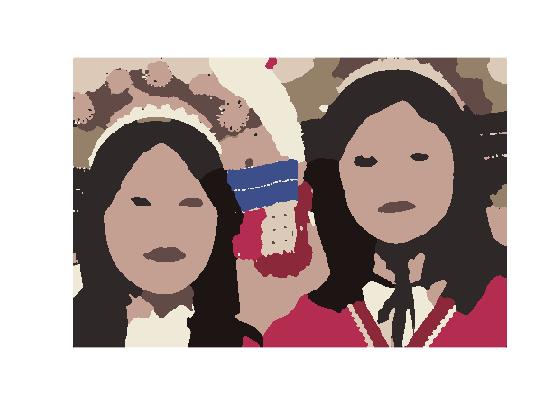}\label{lady:l2}}
\caption{(a): Original image;  (b), (c), (d): Segmentation results obtained from the  Potts model with the region force functions~ \eqref{eq:region_force1},  \eqref{eq:region_force2}, and  \eqref{eq:region_L2}, respectively.}
\label{fig:lady}
\end{figure*}

\begin{figure*}[!ht]
\centering
\subfloat[]{\includegraphics[width=0.5\textwidth]{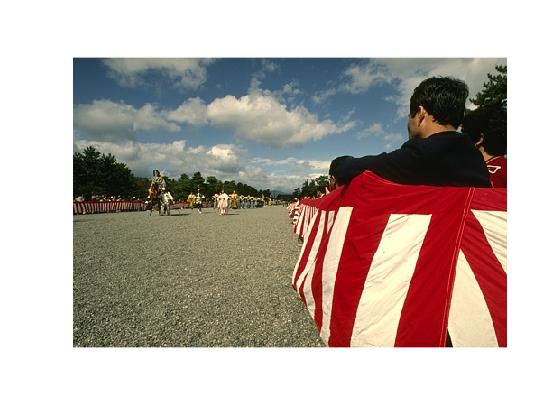}\label{flag:orig}} \hfil
\subfloat[]{\includegraphics[width=0.5\textwidth]{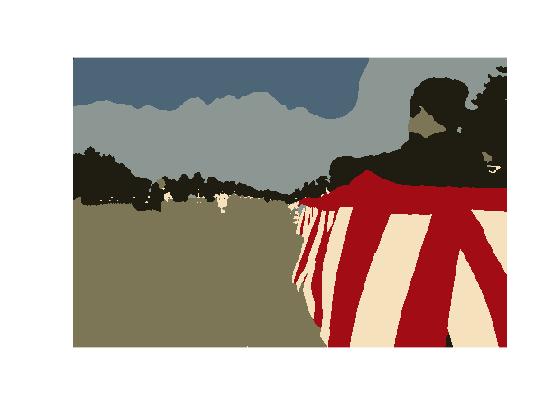}\label{flag:logp}}\\
\subfloat[]{\includegraphics[width=0.5\textwidth]{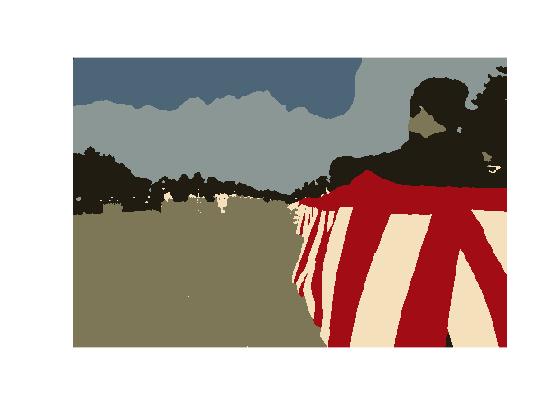}\label{flag:p}}\hfil
\subfloat[]{\includegraphics[width=0.5\textwidth]{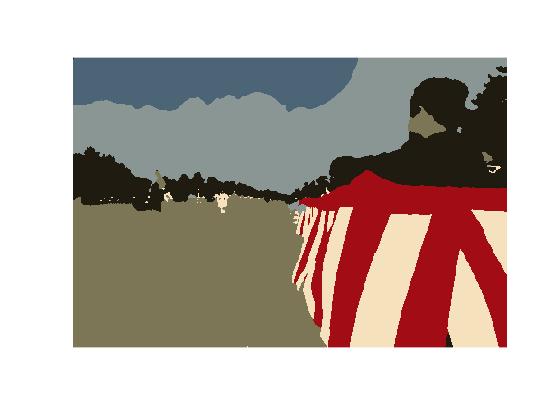}\label{flag:l2}}
\caption{(a): Original image;  (b), (c), (d): Segmentation results obtained from the  Potts model with the region force functions~ \eqref{eq:region_force1},  \eqref{eq:region_force2}, and  \eqref{eq:region_L2}, respectively.}
\label{fig:flag}
\end{figure*}

We first examine the performance of the proposed region force function listed in \eqref{eq:region_force1}  on RGB image segmentation problems, and compare them with the widely used $l_2$ region force function 
\begin{align*}
f_k(x)=|I(x)-c_k|^2\numberthis\label{eq:region_L2},
\end{align*} 
and the one in \eqref{eq:region_force2}. The test images  are obtained from BSDS500 \cite{Malik2004bsds}, see Figures~\ref{chapel:orig},~\ref{veg:orig}, \ref{lady:orig}, and \ref{flag:orig}. They are discretized on a Cartesian grid with the weights on all the edges being equal to $1$, and the image density centroids  (i.e., $c_k$, $1\leq k\leq K$) are  computed by the {\tt{kmeans}} algorithm in Matlab. We use  un-smoothed images (i.e., $\sigma=0$) to compute the edge detector in \eqref{eq:edge_detector}. 
The values of $K$ and the values of $\beta,~\gamma$   in each image segmentation test are listed in Table~\ref{tab:img_par}. 
 The probability function $p_k(x)$ in \eqref{eq:region_force1} and \eqref{eq:region_force2} is computed via \eqref{eq:image_prob} with  unit variance. 
 
\begin{table}[!ht]
\caption{Parameters used for the image segmentation tests in Figures~\ref{chapel:logp} to \ref{chapel:l2}, \ref{veg:logp} to \ref{veg:l2}, \ref{lady:logp} to \ref{lady:l2}, and \ref{flag:logp} to \ref{flag:l2}.}\label{tab:img_par}
\centering
\begin{tabular}{|c|c||c|c|c||c|c|c|}
\hline
& \ref{chapel:logp} & \ref{chapel:p} &\ref{chapel:l2} &\ref{veg:logp} & \ref{veg:p} &\ref{veg:l2}\\\hline
 K & 4& 4&4 & 7& 7&7\\\hline
$\beta$ & 0.6& 0.3&0.5& 0.6& 0.25& 0.5\\\hline
 $\gamma$ & 50& 70&70 & 55& 75& 60\\\hline\hline
 & \ref{lady:logp} & \ref{lady:p} &\ref{lady:l2} &\ref{flag:logp} & \ref{flag:p} &\ref{flag:l2}\\\hline
 K & 10& 10&10 & 6&6& 6\\\hline
$\beta$ & 1.35& 0.45&1.35 &1.45& 0.5& 1.35\\\hline
 $\gamma$ & 55& 100&55 & 45& 55& 55\\\hline
\end{tabular}
\end{table}


We solve the Potts model \eqref{eq:potts_model_conv} with the three different region force functions using PDHG (Algorithm~\ref{alg:pdhg}) and ADMM  (Algorithm~\ref{alg:admm}), where we set $\beta_l=\gamma_l=0.4$  in PDHG, and $\beta_l=0.05$ and $c=0.1$ in ADMM. In order to monitor the convergence of the algorithm, we record two quantities: The primal energy 
\begin{equation}\label{eq:primal_energy}
E_P([\phi_k])= \int_{\Omega}\sum_{i=1}^{K}f_{k} \phi_{k}+\alpha|\nabla\phi_{k}|\,\mathrm{d}x,
\end{equation}
and the dual energy
\begin{equation}
E_D([q_k]) = \int_{\Omega}\underset{k\in\left\{ 1,\ldots,K\right\} }{\min}(f_{k}+\mathrm{div}q_{k})\,\mathrm{d}x.
\end{equation}
Because the primal variables $[\phi_k]$ from Algorithm~\ref{alg:admm} could violate the simplex constraint, the duality gap is not always great than or equal to zero. However, as the algorithm converges, the duality gap is approaching zero. Therefore, both PDHG and ADMM are terminated if either of the following two conditions is satisfied:
a) a maximum of $2500$ iterations is reached;
b) the relative absolute duality gap is smaller than $\epsilon$,
\begin{align*}
\frac{|E_P-E_D|}{|E_P|}\leq \epsilon,\numberthis\label{eq:stopping_epsilon}
\end{align*}
where $\epsilon=10^{-5}$ in the experiments for image segmentation.

The segmentation results obtained from PDHG and ADMM are visually close, so we only present the ones obtained from PDHG in Figures~\ref{fig:chapel} to \ref{fig:flag}. 
These figures show that for image segmentation the new region force function \eqref{eq:region_force1} proposed in this paper  is as effective as the  $L_2$ fidelity and the one proposed in \cite{yin2016newregion}. We direct the interested reader to \cite{DRK2015PAMI} for indirect comparisons of the segmentation results with other approaches.

We present the computational results of PDHG and ADMM in Table~\ref{tab:timing}, which shows both methods can solve the segmentation with very good efficiency. For these test images, we find that PDHG requires many fewer iterations and less computation time than ADMM to converge to the moderate accuracy. However, the computing time could also be in favor of the ADMM method for some other images. The table also shows an interesting feature about the three region force functions. Typically, it  requires the least number of iterations and computation time for PDHG and ADMM to compute the solution of the Potts model with the region force function \eqref{eq:region_force1}, while it requires the most number of iterations and computation time to compute the solution of the Potts model with the region force function \eqref{eq:region_L2}.
\begin{table}[!ht]
\caption{Number of iterations and computational time of PDHG and ADMM for the image segmentation tests in Figures~\ref{chapel:logp} to \ref{chapel:l2}, \ref{veg:logp} to \ref{veg:l2}, \ref{lady:logp} to \ref{lady:l2}, and \ref{flag:logp} to \ref{flag:l2}.}\label{tab:timing}
\centering
\begin{tabular}{|c|c||c|c|c||c|c|c|}
\hline
& & \ref{chapel:logp} & \ref{chapel:p} &\ref{chapel:l2} &\ref{veg:logp} & \ref{veg:p} &\ref{veg:l2}\\\hline
\multirow{ 2}{*}{ADMM} & \#iter & 352&1033 &2500 & 921& 2500& 2500 \\\cline{2-8}
& time (s) & 162.8& 417.5&1144.2 & 761.3& 2062.8& 2043.5\\\hline
\multirow{ 2}{*}{PDHG} & \#iter & 176& 307& 642& 826& 670& 2500 \\\cline{2-8}
& time (s) & 58.9& 108.9& 224.1& 492.3& 432.1& 1344.8\\\hline\hline
& & \ref{lady:logp} & \ref{lady:p} &\ref{lady:l2} &\ref{flag:logp} & \ref{flag:p} &\ref{flag:l2}\\\hline
\multirow{ 2}{*}{ADMM} & \#iter & 1911& 2500&2500 & 2363& 2500&2500  \\\cline{2-8}
& time (s) & 2231.6& 2922.9& 2930.5& 1671.2& 1751.8&1754.6 \\\hline
\multirow{ 2}{*}{PDHG} & \#iter & 1751& 2500&2500 & 744& 562&2500  \\\cline{2-8}
& time (s) & 1392.9& 1905.1& 1895.7& 386.9& 305.4&1151.3 \\\hline

\end{tabular}
\end{table}

\subsection{Semi-supervised data clustering}
Next, we evaluate the performance of the new region force function on three benchmark semi-supervised clustering data sets:  \texttt{Three-Circles}, \texttt{COIL}, and \texttt{MNIST}.  \texttt{Three-Circles} is a synthetic data set which are constructed  from three circles having an identical center. We first create three circles on the 2D plane, centered at (0,0) with radii 1, 2, and 3, and then sample 6000 points  uniformly at random from these circles. The sampled points are embedded into $\mathbb{R}^{100}$ by padding 98 zeros to their  end, followed by the perturbation of each coordinate  with i.i.d Gaussian noise of mean $0$ and variance $0.16$. 
The \texttt{COIL} data set is downloaded from the supplementary material of \cite{ChaSchZie06} (\url{http://olivier.chapelle.cc/ssl-book/benchmarks.html}, originally from COIL-100 \cite{COIL100}). It contains 1500 natural images of 6 different objects taken from various angles. All the images are preprocessed to the same size, and the labels for the images are also contained in the data set.  \texttt{MNIST} is obtained from ``The MNIST Database of Handwritten Digits'' (\url{http://yann.lecun.com/exdb/mnist/}), which consists of 70,000 gray-scale images of labeled handwritten digits from 0 to 9, all scaled to the same size.  The basic properties of the three data sets are listed in Table \ref{tab:bench}.

\begin{table}[h]
\caption{Basic properties of \texttt{Three-Circles}, \texttt{COIL}, and \texttt{MNIST}. The original data sets
also contain labels for all the data points}\label{tab:bench}
\centering
\begin{tabular}{|c||c|c|c|}
\hline
\textbf{Data set} & \textbf{Classes} & \textbf{Dimension} & \textbf{Points}\\
\hline
\texttt{Three Circles} & 3 & 100 & 6000\\
\hline
\texttt{COIL} & 6 & 241 & 1500\\
\hline
\texttt{MNIST} & 10 & 784 & 70,000\\
\hline
\end{tabular}
\end{table}

The graph $G$ for each test data set is constructed as a  $s$-nearest-neighbor ($s$-NN) graph under the $l_2$-metric.  
We make use of an implementation of the randomized kd-tree \cite{silpa-anan_optimised_2008, muja_fast_2009}, called {\tt{VLFeat}}  \cite{vedaldi08vlfeat}, to find the $s$-nearest neighbors of each data point. The Zelnik-Manor and Perona weight function in \eqref{eq:zmp_kernel}  is used to construct the affinity matrix, where the standard deviation at a data point is estimated using the distance between the data point and its $s$-th nearest neighbor. 

In the tests, a small fraction of data points are drawn  uniformly at random  from each data set and marked  as labelled data, and then we apply the discrete Potts model \eqref{eq:potts_discrete_conv} to determine the labels for the rest of the data points.
The probabilities $p_k(x_i)$ used to define the region force functions \eqref{eq:region_force1} and \eqref{eq:region_force2} are computed via \eqref{eq:region_probability} for $m=1$ or $m=2$. We choose $\alpha(x_i)$ to be a constant, denoted by $\alpha$, in \eqref{eq:potts_discrete_conv} . The solution to the discrete Potts model is computed by PDHG (Algorithm~\ref{alg:pdhg}) and ADMM (Algorithm~\ref{alg:admm}). The algorithms are terminated using the same criteria as in Section~\ref{sec:num_image} but with $\epsilon=10^{-3}$. All the simulations are repeated $10$ times.

\begin{table}[!ht]
\caption{Parameters used in the tests where $n$ is the number of labelled data points in each data set, $s$ is the number of neighbors used to construct the graph, $m$ is the value used in the computation of $p_k(x_i)$, and $\alpha$ is the TV weight in the Potts model. The last two rows include the parameters used in PDHG and ADMM.}\label{tab:clustering_para}
\centering
\begin{tabular}{|c||c|c|c|}
\hline
& \texttt{Three Circles} &\texttt{COIL} &\texttt{MNIST}\\\hline
$n$ & 50 &100 &350\\\hline
$s$ & 10&5 &10 \\\hline
$m$ &2 &1 &2 \\\hline
$\alpha$ for \eqref{eq:region_force1} &3 &5.5 &5.5 \\\hline
$\alpha$ for \eqref{eq:region_force2} &0.5 &1.5 &1.5 \\\hline
PDHG &$\beta_l=\gamma_l=0.4$ &$\beta_l=\gamma_l=0.4$ &$\beta_l=0.5l$, $\gamma_l=\frac{0.5}{(1+0.1l)}$  \\\hline
ADMM &$\beta_l=0.05$, $c=0.05$ &$\beta_l=0.05$, $c=0.1$ & $\beta_l=0.05$, $c=5$\\\hline
\end{tabular}
\end{table}

\begin{figure}[!ht]
\centering
\subfloat[]{\label{fig:3circles_sample}\includegraphics[width=0.33\linewidth]{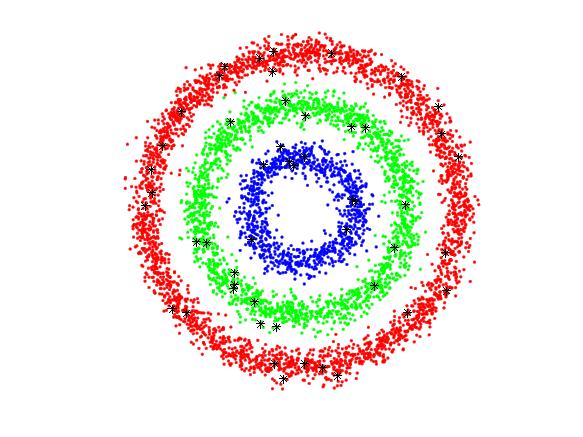}}
\subfloat[]{\label{fig:p_k_2}\includegraphics[width=0.33\linewidth]{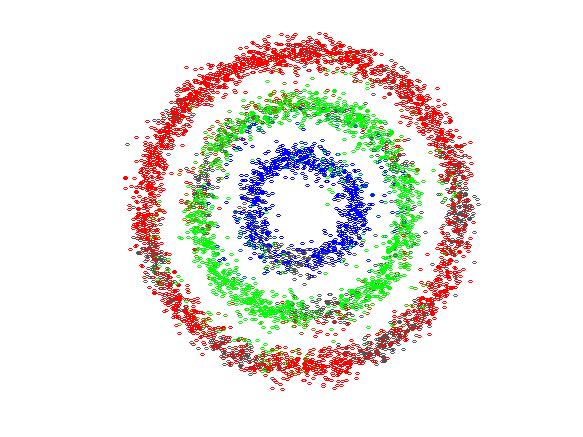}}
\subfloat[]{\label{fig:I_k}\includegraphics[width=0.33\linewidth]{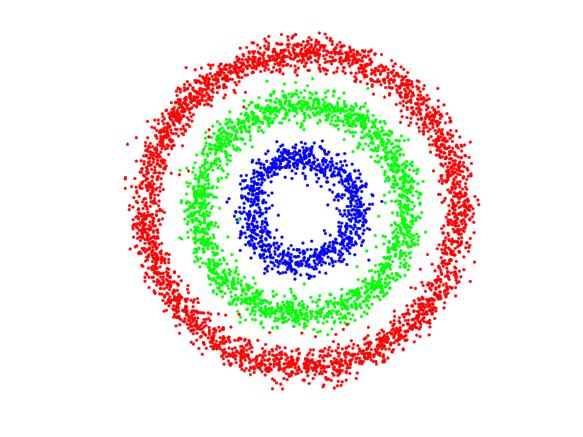}}
\caption{\texttt{Three-Circle} synthetic data: 
(a) all the data points with labelled data, (b) classification result computed from $p_k(x_i)$, and (c) classification result of the Potts model with \eqref{eq:region_force1}.}
\label{fig:3moons}
\end{figure}

We begin with comparing the performance of the region force functions \eqref{eq:region_force1} and \eqref{eq:region_force2}. When the region force function \eqref{eq:region_force1} is used in the discrete Potts model, we add a small quantity $\delta=10^{-3}$ to the logarithm and test 
\begin{align*}
f_k(x) =-\log(p_k(x)+\delta)+ \log(1-p_k(x)+\delta)
\end{align*}
to avoid numerical overflow. The values of the parameters used in our tests for the three different data sets are listed in Table~\ref{tab:clustering_para}. 

The average classification accuracy (out of the ten random samples of the labelled data) of the Potts model with the region force functions \eqref{eq:region_force1} and \eqref{eq:region_force2} are listed in Tables~\ref{tab:3moon_result}, \ref{tab:coil_result}, and \ref{tab:mnist_result} for {\tt Three Circles}, {\tt  COIL}, and {\tt MNIST}, respectively. 
The three tables show that the average classification accuracy of  \eqref{eq:region_force1} is slightly higher than that of \eqref{eq:region_force2}, and overall they are equally effective for the tested semi-supervised clustering problems. 
We also plot the classification result of the Potts model with \eqref{eq:region_force1} for {\tt Three Circles} in Figure~\ref{fig:3moons} by projecting the data points onto the first two dimensions. The  average computation time  and average number of iterations of PDHG and ADMM in each test are also listed in the tables, which show that PDHG is typically faster than ADMM.

\begin{table}
[!ht]
\caption{Average classification accuracy of the discrete Potts model with the region forces functions \eqref{eq:region_force1} and \eqref{eq:region_force2} for {\tt Three-Circles}, as well as the average computational time of PDHG and ADMM. The number of labelled data points is 50 (0.83\%).}\label{tab:3moon_result}
\centering
\begin{tabular}{|c||c|c|c|c|}
\hline
   region force & algorithm & accuracy (\%) & iterations (ave.)  & cpu time (s) \\
  \hline
 \eqref{eq:region_force1} & PDHG & 98.2  & 162.8  & 2.94    \\
 \hline
 \eqref{eq:region_force1} & ADMM  & 98.2  & 76.3  & 2.78 \\\hline
 \eqref{eq:region_force2} & PDHG & 97.9 & 65.6  &  1.45  \\
 \hline
 \eqref{eq:region_force2} & ADMM& 97.9  &  71.5 & 2.70    \\
 \hline
\end{tabular}
\end{table}

\begin{table}
[!ht]
\caption{Average classification accuracy of the discrete Potts model with the region forces functions \eqref{eq:region_force1} and \eqref{eq:region_force2} for {\tt COIL}, as well as the average computational time of PDHG and ADMM. The number of labelled data points is 100 (6.7\%).}\label{tab:coil_result}
\centering
\begin{tabular}{|c||c|c|c|c|}
\hline
   region force & algorithm & accuracy (\%) & iterations (ave.)  & cpu time (s) \\
 \hline
  \eqref{eq:region_force1} & PDHG & 90.90 & 307.6 & 2.16  \\
 \hline
 \eqref{eq:region_force1} & ADMM  & 90.90 & 163.1 & 1.67 \\
 \hline
 \eqref{eq:region_force2} & PDHG & 90.25 & 306.6 & 2.09  \\
 \hline
 \eqref{eq:region_force2} & ADMM& 90.25 & 475.9 & 4.21  \\
 \hline
\end{tabular}
\end{table}

\begin{table}
[!ht]
\caption{Average classification accuracy of the discrete Potts model with the region forces functions \eqref{eq:region_force1} and \eqref{eq:region_force2} for {\tt MNIST}, as well as the average computational time of PDHG and ADMM. The number of labelled data points is 350 (0.5\%).}\label{tab:mnist_result}
\centering
\begin{tabular}{|c||c|c|c|c|}
\hline
   region force & algorithm & accuracy (\%) & iterations (ave.)  & cpu time (s) \\
 \hline
 \eqref{eq:region_force1} & PDHG & 97.3 & 110.1   &  81.7   \\
 \hline
 \eqref{eq:region_force1} & ADMM & 97.3  & 385.8   & 2097   \\
 \hline
 \eqref{eq:region_force2} & PDHG & 97.2    & 203.8 &  161.3  \\
 \hline
 \eqref{eq:region_force2} & ADMM& 97.2   & 381.6   & 2063    \\
 \hline
 
\end{tabular}
\end{table}

We further compare our approach, referred to as Potts-RF, with another two existing variational methods from the literature:  multiclass total variation (MTV \cite{bresson2013multiclass}) and multiclass-MBO \cite{garcia-cardona_multiclass_2014}.
The codes for MTV are downloaded from the author's website, while we reproduce the codes for multiclass MBO using the parameters suggested in \cite{garcia-cardona_multiclass_2014}. We test three different numbers of labeled samples for each data set. The average classification accuracy of Potts-RF, MTV and multiclass-MBO is listed in Tables~\ref{tab:threemoon_result2}, \ref{tab:COIL_result2}, and \ref{tab:mnist_result2} for {\tt Three Circles}, {\tt  COIL}, and {\tt MNIST}, respectively.  Table~\ref{tab:mnist_result2} shows that the classification accuracy of Potts-RF is only about 0.5\% lower than that of MTV for {\tt MNIST}, while Tables~\ref{tab:threemoon_result2} and \ref{tab:COIL_result2} show that the classification accuracy of Potts-RF is larger than that of MTV and multi-class MBO for the other two data sets. 
In addition, our approach is much easier to be implemented than MTV and multi-class MBO which requires  either complicated initialization or computation of the eigenvectors of a large matrix. 
For the sake of completeness, we also include the classification accuracy computed from the initial probabilities $p_k(x_i)$ in the tables.
\begin{table} 
[!ht]
\caption{Average classification accuracy (\%) of Potts-RF, MTV and multiclass-MBO on \texttt{Three-Circles} for three different numbers of labeled samples.}  \label{tab:threemoon_result2}
\centering
\begin{tabular}{|c||c|c|c|}
\hline
$l$ & 0.83\% & 1.25\% & 1.67\% \\ 
 \hline
  $p_k(x_i)$ & $75.92 \pm 2.43$ & $83.42\pm 2.45$ & $89.91 \pm 1.27$ \\
 \hline
Potts-RF\eqref{eq:region_force1}   & $98.19\pm 3.58$ & $99.35\pm 0.07$   & $99.49\pm0.05$   \\
 \hline
 MTV & $75.44 \pm 8.34$  & $79.19\pm4.96$ & $79.47\pm 2.03$ \\
 \hline
 multiclass-MBO & $66.15\pm 5.98$   & $81.13 \pm 5.53$   & $90.02\pm 3.31$  \\
 \hline
\end{tabular}
\end{table}

\begin{table}
[!ht]
\caption{Average classification accuracy (\%) of Potts-RF, MTV and multiclass-MBO on \texttt{COIL} for three different numbers of labeled samples.}  \label{tab:COIL_result2}
\centering
\begin{tabular}{|c||c|c|c|}
\hline
 & 3.3\% & 6.7\% &10\% \\
 \hline
 $p_k(x_i)$ & $46.64 \pm 1.79$  & $58.90\pm 2.41$ & $ 66.46\pm 2.31$ \\
 \hline
Potts-RF\eqref{eq:region_force1} & {${81.8 \pm 4.9}$} & ${ 90.9 \pm 2.0}$ & ${ 92.9 \pm 0.9}$ \\
 \hline
MTV & $78.4 \pm 4.00$  & $89.73\pm1.5$   & $92.20\pm 1.3$   \\
 \hline
 multiclass-MBO & $70.53\pm3.46$  & $82.03\pm 3.90$  & $89.09\pm 2.06$   \\
 \hline
\end{tabular}
\end{table}

\begin{table}
[!ht]
\caption{Average classification accuracy (\%) of Potts-RF, MTV and multiclass-MBO on \texttt{MNIST} for  three different numbers of labeled samples.}  \label{tab:mnist_result2}
\centering 
\begin{tabular}{|c||c|c|c|}
\hline
 & 0.25\% & 0.5\% &1\% \\
  \hline
 $p_k(x_i)$ & $18.24\pm 3.34$ & $24.67\pm 0.99$ & $35.85\pm 0.77$ \\
 \hline
Potts-RF\eqref{eq:region_force1}   & $97.15 \pm 0.13$  & $97.28 \pm 0.09$ & $97.32\pm0.09$  \\
 \hline
 MTV & $97.62 \pm 0.03$  &  $97.63 \pm 0.03$ & $97.65 \pm 0.01$ \\
 \hline
 multiclass-MBO & $73.0\pm 3.91$  & $90.1\pm 3.24$  & $94.9\pm 2.78$   \\
 \hline
\end{tabular}
\end{table}

\section{Conclusion and Future Direction}\label{sec:con}
We introduce a novel region force function into the Potts model and thus provide a uniformly effective framework for multi-phase image segmentation and semi-supervised data clustering. The new region force function is computed as the negative log-likelihood function of the indicator function under the Bernoulli distribution. The probability that an image pixel or a data point belongs to a given class is  estimated based on the mixed Gaussian density model for image segmentation and based on the diffusion distance for semi-supervised data clustering.

Two numerical algorithms PDHG and ADMM are presented to compute the solution of the Potts model. Those two algorithms are developed from two different dual formulations of the Potts model. Extensive numerical experiments have been conducted on benchmark problems in image segmentation and semi-supervised data clustering and show that our approach is as effective as other existing variational methods in the literature.

In this paper, the probabilities used in the computation of the  region force function are fixed. For future work, we suggest updating the probabilities adaptively in the numerical algorithms, for example based on the maximum likelihood estimation. We also intend to apply data driven ideas to  design  new region force functions for different applications.
\bibliographystyle{acmtrans-ims}
\bibliography{MaxFlow}
\end{document}